\documentclass{article} 
\usepackage{iclr2024_conference,times}

\usepackage{amsmath,amsfonts,bm}

\def\eqref#1{equation~\ref{#1}}

\def\1{\bm{1}}

\def\vu{{\bm{u}}}

\def\vx{{\bm{x}}}

\DeclareMathAlphabet{\mathsfit}{\encodingdefault}{\sfdefault}{m}{sl}
\SetMathAlphabet{\mathsfit}{bold}{\encodingdefault}{\sfdefault}{bx}{n}

\def\sX{{\mathbb{X}}}

\usepackage{hyperref}
\usepackage{url}

\usepackage{amsmath,amssymb,amsfonts}
\usepackage{algorithmic}
\usepackage{graphicx}
\usepackage{textcomp}
\usepackage{xcolor}
\usepackage{xspace} 
\usepackage{hyperref}
\usepackage{multirow}
\usepackage{floatrow}
\usepackage{booktabs}

\usepackage{caption}
\usepackage{subcaption}

\usepackage{float}
\floatstyle{plaintop}
\restylefloat{table}
\newtheorem{proposition}{Proposition}
\newtheorem{theorem}{Theorem}

\usepackage{cleveref}

\title{Formatting Instructions for ICLR 2024 \\ Conference Submissions}

\iclrfinalcopy 
\begin{document}

\title{GPS-SSL: Guided Positive Sampling to Inject Prior Into Self-Supervised Learning 
}

\author{Aarash Feizi$^{1, 2}$, Randall Balestriero$^3$, Adriana Romero-Soriano$^{1, 2}$, Reihaneh Rabbany$^{1, 2}$ \\ \\
$^1$McGill University, $^2$Mila Institute, $^3$ Independent\\
\texttt{\{aarash.feizi, rrabba\}@mail.mcgill.ca}\\
\texttt{\{randallbalestriero, adriana.romsor\}@gmail.com} \\
}

\definecolor{myred}{RGB}{204, 0, 0}
\definecolor{mygreen}{RGB}{0, 99, 0}
\definecolor{myblue}{RGB}{0, 0, 99}
\definecolor{myyellow}{RGB}{204, 204, 0}

\newcommand{\cut}[1]{}
\newcommand{\firstval}{\text{$\mathcal{D}^{val}_{SS}$}\xspace}
\newcommand{\secondval}{\text{$\mathcal{D}^{val}_{SU}$}\xspace}
\newcommand{\thirdval}{\text{$\mathcal{D}^{val}_{UU}$}\xspace}
\newcommand{\fourthval}{\text{$\mathcal{D}^{val}_{??}$}\xspace}

\newcommand{\firsttest}{\text{$\mathcal{D}_{SS}$}\xspace}
\newcommand{\secondtest}{\text{$\mathcal{D}_{SU}$}\xspace}
\newcommand{\thirdtest}{\text{$\mathcal{D}_{UU}$}\xspace}
\newcommand{\fourthtest}{\text{$\mathcal{D}_{??}$}\xspace}

\newcommand{\hotelsfifty}{Revisited Hotels-50K\xspace}
\newcommand{\hotelsid}{R-HID\xspace}
\newcommand{\hotelsidfullname}{Revised-Hotel-ID\xspace}

\newcommand{\randauroc}{$\text{AUC}$\xspace}

\newcommand{\newauroc}{$\text{AUC}_{H}$\xspace}
\newcommand{\recallatone}{R@1\xspace}

\newcommand{\s}[2]{s(#1, #2)}

\newcommand{\ourmethod}{\text{GPS-SimCLR}\xspace}

\newcommand{\ourstrategy}{\text{GPS-SSL}\xspace}
\newcommand{\oursimclr}{\text{GPS-SimCLR}\xspace}
\newcommand{\ourbyol}{\text{GPS-BYOL}\xspace}
\newcommand{\ourbarlow}{\text{GPS-Barlow}\xspace}
\newcommand{\ourvicreg}{\text{GPS-VICReg}\xspace}

\newcommand{\ourmethodfullname}{\text{GPS-SimCLR}\xspace}
\newcommand{\ourstrategyfullname}{\text{Guided Positive Sampling Self-Supervised Learning}\xspace}

\newcommand{\weakaug}{\text{\textit{RHFlipAug}}\xspace}
\newcommand{\strongaug}{\text{\textit{StrongAug}}\xspace}

\newcommand{\weak}{\text{\textit{RHFlip}}\xspace}
\newcommand{\strong}{\text{\textit{Strong}}\xspace}

\newcommand{\aircrafts}{\text{FGVCAircraft}\xspace}
\newcommand{\cifarten}{\text{Cifar10}\xspace}
\newcommand{\cifarhundred}{\text{Cifar100}\xspace}
\newcommand{\svhn}{\text{SVHN}\xspace}
\newcommand{\pathmnist}{\text{PathMNIST}\xspace}
\newcommand{\tissuemnist}{\text{TissueMNIST}\xspace}
\newcommand{\kk}{$k$\xspace} 
\newcommand{\NNkj}{$NN_k^j$\xspace} 
\newcommand{\Ij}{$I^j$\xspace} 

\newcommand{\FB}[1]{\text{\textbf{#1}}} 

\newcommand{\SB}[1]{\textcolor{blue}{#1}} 
\newcommand{\TB}[1]{\textcolor{red}{#1}} 

\newcommand{\edit}[1]{#1}

\maketitle

\begin{abstract}

We propose \textit{\ourstrategyfullname} (\ourstrategy), a general method to inject a priori knowledge into Self-Supervised Learning (SSL) positive samples selection. 
Current SSL methods leverage Data-Augmentations (DA) for generating positive samples and incorporate prior knowledge--an incorrect, or too weak DA will drastically reduce the quality of the learned representation. 
\ourstrategy proposes instead to design a metric space where Euclidean distances become a meaningful proxy for semantic relationship. In that space, it is now possible to generate positive samples from nearest neighbor sampling. Any prior knowledge can now be embedded into that metric space independently from the employed DA. From its simplicity, \ourstrategy is applicable to any SSL method, e.g. SimCLR or BYOL.
A key benefit of \ourstrategy is in reducing the pressure in tailoring strong DAs. For example \ourstrategy reaches 85.58\% on \cifarten with weak DA while the baseline only reaches 37.51\%.  
We therefore move a step forward towards the goal of making SSL less reliant on DA.
We also show that even when using strong DAs, \ourstrategy outperforms the baselines on under-studied domains.
We evaluate \ourstrategy along with multiple baseline SSL methods on numerous downstream datasets from different domains when the models use \textit{strong} or \textit{minimal} data augmentations.
We hope that \ourstrategy will open new avenues in studying how to inject a priori knowledge into SSL in a principled manner.
\end{abstract}


\section{Introduction}
\label{sec:intro}



Self-supervised learning (SSL) has recently shown to be one of the most effective learning paradigm across many data domains~\citep{radford2021learning, girdhar2023imagebind, assran2023self, chen2020simple, grill2020bootstrap, bardes2021vicreg, balestriero2023cookbook}. 
SSL belongs to the broad category of annotation-free representation learning approaches, which have enabled machine learning models to use abundant and easy-to-collect unlabeled data, facilitating the training of ever-growing deep neural network architectures.\looseness-1 

Despite the SSL promise, current approaches require handcrafted a priori knowledge to learn useful representations. This a priori knowledge is often injected through the positive sample--\textit{i.e.}, semantically related samples--generation strategies employed by SSL methods~\citep{chen2020simple}. In fact, SSL representations are learned so that such positive samples get as similar as possible in embedding space, all while preventing a collapse of the representation to simply predicting a constant for all inputs. 
The different strategies to achieve that goal lead to different flavors of SSL methods~\citep{chen2020simple, grill2020bootstrap, bardes2021vicreg, zbontar2021barlow, chen2021exploring}.
In computer vision, positive sample generation mostly involves sampling an image from the dataset, and applying multiple handcrafted and heavily tuned data augmentations (DAs) to it, such as rotations and random crops, which preserve the main content of the image.
\looseness-1

The importance of selecting the right DAs is enormous as it impacts performances to the point of producing a near random representation, in the worst case scenario~\citep{balestriero2023cookbook}. As such, tremendous time and resources have been devoted to designing optimal DA recipes, most notably for eponymous datasets such as ImageNet~\citep{deng2009imagenet}.
From a practitioner's standpoint, positive sample generation could thus be considered solved if one were to deploy SSL methods \emph{only} on such popular datasets. 
Unfortunately--and as we will thoroughly demonstrate throughout this paper--common DA recipes used in those settings fail to transfer to other datasets. 
For example, since ImageNet consists of natural images mostly focused around 1000 different object categories, we observe a reduction of performance on datasets consisting of more specialized images, such as hotel room images \citep{stylianou2019hotels, kamath20212021}, or only focused on different types of airplanes \citep{maji2013fine}, or medical images \citep{medmnistv2}.
Since searching for the optimal DAs is computationally intense, there remains an important bottleneck when it comes to deploying SSL to new or under-studied datasets. This becomes particularly important when applying SSL methods on data gathered for real-world applications.\looseness-1

In this paper, we introduce a strategy to obtain positive samples, which generalizes the well established NNCLR SSL method~\citep{dwibedi2021little}. 
While NNCLR proposes to obtain positive samples by leveraging known DAs and nearest neighbors in the embedding space of the network being trained, we propose to perform nearest neighbour search in an independently constructed embedding space. One special case could be to employ the embedding space of the network being trained--therefore recovering NNCLR--but more interestingly may also be generated by any pre-trained network, or even fully hand-crafted. This flexibility allows to (i) enable simple injection of prior knowledge into positive sampling without relying on tuning the DA, and most importantly (ii) makes the underlying SSL method much more robust to under-tuned DAs.
By construction, the proposed method coined \ourstrategy for \ourstrategyfullname, can be coupled off-the-shelf with any SSL method used to learn representations, \textit{e.g.}, BarlowTwins~\citep{zbontar2021barlow}, SimCLR~\citep{chen2020simple}, BYOL~\citep{grill2020bootstrap}. We validate the proposed \ourstrategy approach on a benchmark suite of under-studied datasets, namely \aircrafts, \pathmnist, \tissuemnist, and show remarkable improvements over baseline SSL methods. We further evaluate our model on a real-world dataset, \hotelsidfullname (\hotelsid) \citep{feizi2022revisiting} and show clear improvements of our method compared the baseline SSL methods.
Finally, we validate the approach on commonly used image datasets with known effective DAs recipes, and show that GPS remains competitive.
Through comprehensive ablations, we show that \ourstrategy takes a step towards shifting the focus of designing \emph{well-crafted DAs} to having a better \emph{prior knowledge} embedding space in which choosing the nearest neighbour becomes an attractive positive sampling strategy.\looseness-1


The main contributions of this paper can be summarized as follows:
\begin{itemize}
    \item We propose a positive sampling strategy, \ourstrategyfullname (\ourstrategy) that enables SSL models to use prior knowledge about the target-dataset to help with the learning process and reduce the reliance on carefully hand-crafted augmentations. The prior knowledge is an embedding space in which Euclidean based nearest neighbor sampling is informative of semantic relationship.~\looseness-1
    \item We evaluate \ourstrategy by applying it to baseline SSL methods and show that with strong augmentations, they perform comparable to, or better than, the original methods. Moreover, they significantly outperform the original methods when using minimal augmentations, making it suitable for learning under-studied or real-world datasets (rather than transfer-learning).
    \item We further evaluate our model on datasets with under-studied applications, we consider hotel retrieval task in the counter human trafficking domain. Similar to benchmark datasets, we see on this less studied dataset, our proposed \ourstrategy outperforms the baseline SSL methods by a significant margin. 
\end{itemize}
We provide the code for \ourstrategy and downloading and using \hotelsid on GitHub, available at:  \url{https://github.com/aarashfeizi/gps-ssl}, for the research community.

\section{Related Work}
\label{sec:background}


Self Supervised Learning (SSL) is a particular form of unsupervised learning methods in which a given Deep Neural Network (DNN) learns meaningful representations of their inputs without labels. 

The variants of SSL are numerous. At the broader scale, SSL defines a pretext task on the input data and train themselves by solving the defined task. In SSL for computer vision, the pretext tasks generally involve creating different views of images and encoding both so that their embeddings are close to each other. However, that criteria alone would not be sufficient to learning meaningful representations as a degenerate solution is for the DNN to simply collapse all samples to a single embedding vector. As such, one needs to introduce an ``anti-collapse'' term. Different types of solutions have been proposed for this issue, splitting SSL methods into multiple groups, three of which are: 1) Contrastive\citep{chen2020simple, dwibedi2021little, kalantidis2020hard}: this group of SSL methods prevent collapsing by considering all other images in a mini-batch as negative samples for the positive image pair and generally use the InfoNCE \citep{oord2018representation} loss function to push the negative embeddings away from the positive embeddings.  2) Distillation\citep{grill2020bootstrap, he2020momentum, chen2021exploring}: these methods often have an asymmetric pair of encoders, one for each positive view, where one encoder (teacher) is the exponential moving average of the other encoder (student) and the loss only back-propagates through the student encoder. In general, this group prevents collapsing by creating asymmetry in the encoders and defines the pre-text task that the student encoder must predict the teach encoder's output embedding.  3) Feature Decorrelation\citep{bardes2021vicreg, zbontar2021barlow}: These methods focus on the statistics of the embedding features generated by the encoders and defines a loss function to encourage the embeddings to have certain statistical features. By doing so, they explicitly force the generated embeddings not to collapse. For example, \cite{bardes2021vicreg} encourages the features in the embeddings to have high variance, while being invariant to the augmentations and also having a low covariance among different features in the embeddings. Besides these groups, there are multiple other techniques for preventing collapsing, such as clustering methods \citep{caron2020unsupervised, xie2016unsupervised}, gradient analysis methods \citep{tao2022exploring}.~\looseness-1

Although the techniques used for preventing collapse may differ among these groups of methods, they generally require the data augmentations to be chosen and tuned carefully in order to achieve high predictive performance \citep{chen2020simple}. Although choosing the optimal data augmentations and hyper-parameters may be considered a solved problem for popular datasets such as Cifar10 \citep{krizhevsky2009learning} or ImageNet \citep{deng2009imagenet}, the SSL dependency on DA remains their main limitation to be applied to large real-world datasets that are not akin natural images. Due to the importance of DA upon the DNN's representation quality, a few studies have attempted mitigation strategies. For example,  \cite{pmlr-v202-cabannes23a} ties the impact of DA with the implicit prior of the DNN's architecture, suggesting that informed architecture may reduce the need for well designed DA although no practical answer was provided. \cite{cabannes2023active} proposed to remove the need for DA at the cost of requiring an oracle to sample the positive samples from the original training set. Although not practical, this study brings a path to train SSL without DA.
Additionally, a key limitation with DA lies in the need to be implemented and fast to produce. In fact, the strong DA strategies required by SSL are one of the main computational time bottleneck of current training pipelines \citep{bordes2023democratizing}. Lastly, the over-reliance on DA may have serious fairness implications since, albeit in a supervised setting, DA was shown to impact the DNN's learned representation in favor of specific classes in the dataset \citep{NEURIPS2022_f73c0453}.

All in all, SSL would greatly benefit from a principled strategy to embed a priori knowledge into generating positive pairs that does not rely on DA. We propose a first step towards such Guided Positive Sampling (GPS) below.~\looseness-1



\begin{figure*}
    \centering
    \includegraphics[width=1.0\textwidth]{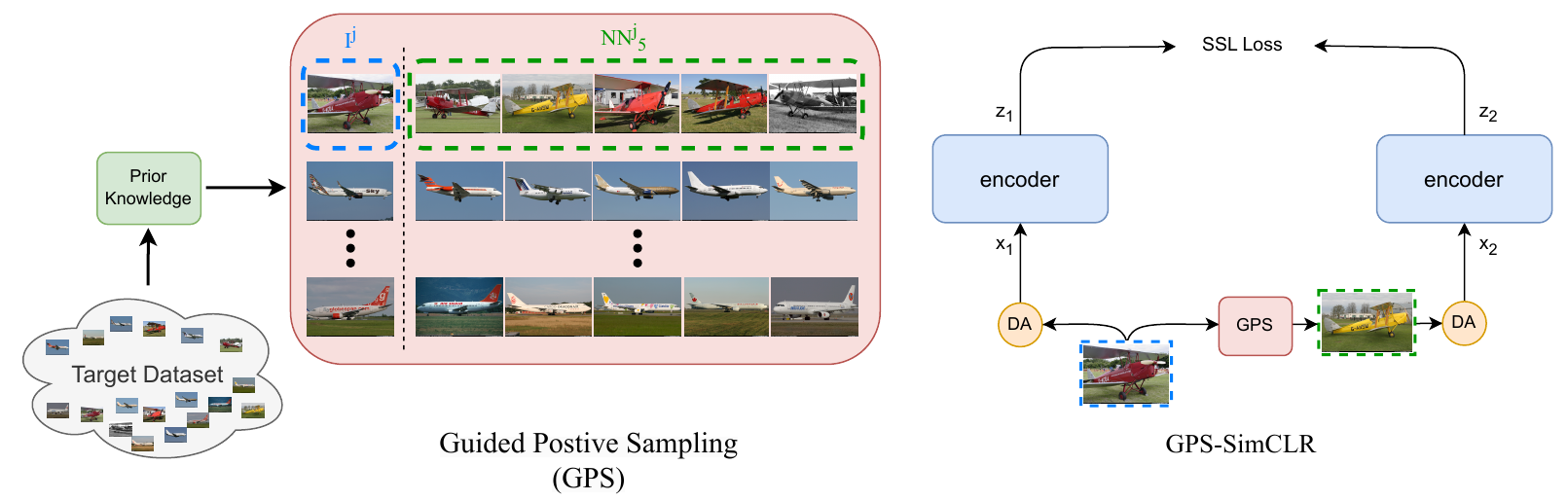}
    \caption{\small Our strategy, \ourstrategy, for positive sampling based on prior knowledge DA-based methods.}
    \label{fig:pknn}
\end{figure*}
\vspace{-2ex}
\section{Guided Positive Sampling for Self-Supervised Learning}
\label{sec:method}
\vspace{-1ex}
We propose a novel strategy, \textit{\ourstrategyfullname} (\ourstrategy), that obtains positive samples by nearest neighbor sampling from a designed embedding space and that greatly reduces the reliance of SSL on well-crafted DA to learning highly informative representations.

\subsection{\ourstrategy: Nearest Neighbor Positive Sampling in a Designed Embedding Space}

As theoretically shown in various studies \citep{haochen2021provable,balestriero2022contrastive,kiani2022joint}, the principal factors that impacts the quality of the learned representation resides in how the positive pairs are defined. In fact, we recall that in all generality, SSL losses that are minimized can mostly be expressed as 
\begin{align}
    \mathcal{L}_{\rm SSL} = \sum_{(\vx,\vx') \in \text{PositivePairs}}\text{Distance}(f_{\theta}(\vx),f_{\theta}(\vx')) - \text{Diversity}(\{f_{\theta}(\vx), \vx \in \sX\}),\label{eq:SSL}
\end{align}
for the mini-batch $\sX$ that gathers all the positive pair samples. A distance measure such as the $\ell_2$ norm or the cosine similarity, and a diversity measure such that the rank of the embedings or proxies of their entropy are used to match the positive pairs embedding all while preventing a representation collapse. All in all, defining the right set of $\text{PositivePairs}$ is what determines the ability of the final representation to solve downstream tasks. The common solution is to repeatedly apply a DA onto a single datum to generate such positive pairs:
\begin{align}
    \text{PositivePairs} \triangleq \{(\text{DA}(\vx),\text{DA}(\vx)), \forall \vx \in \sX\},\label{eq:pospair}
\end{align}
where the $\text{DA}$ operator includes the random realisation of the DA such as the amount of rotation or zoom being applied onto its input image.
However, that strategy often reaches its limits since DAs need to be implemented for the specific data being used. When considering an image dataset, the challenge of designing DA for less common datasets, e.g., FGVCAircraft, led practitioners to instead train the model on a dataset such as ImageNet, where strong DAs have already been discovered, and then transferring the model to other datasets. This however has its limits, e.g. when considering medical images.

As an alternative, we propose an off-the-shelf strategy to sample positive pairs that can be equipped onto any baseline SSL method, e.g., SimCLR, VICReg, coined \ourstrategy and which is defined by defining positive pairs through nearest neighbour sampling in a designed embedding space denoted as $g_{\gamma}$.
First, let's define the collection of samples that are less than $\tau > 0$ away from a query sample $\vx \in \sX$ in the designed embedding space as
\begin{align}
    \mathcal{B}(\vx) \triangleq \{\vx' \in \sX : \|g_{\gamma}(\vx)- g_{\gamma}(\vx')\|_2^2 < \tau\}.\label{eq:GPS1}
\end{align}
From \cref{eq:GPS1}, \ourstrategy simply obtains positive pairs by selecting the furthest point from $\vx$ in $\mathcal{B}(\vx)$ as in
\begin{align}
    \text{PositivePairs}_{\rm GPS} \triangleq \{(\text{DA}(\vx),\text{DA}(\vx')), \forall (\vx, \vx') \in \sX^2 : \vx' = \arg\max_{\vu \in \mathcal{B}(\vx)} \|g_{\gamma}(\vu)- g_{\gamma}(\vx)\|_2^2 \},\label{eq:GPS}
\end{align}
We note that while we keep the DA operator in \cref{eq:GPS}, the quality of the learned representations will be shown to be much less sensitivie to its design and strength compared to \cref{eq:SSL}.
From this, we obtain a first direct result below making \ourstrategy recover a powerful existing method known as NNCLR.

\begin{proposition}
\label{prop:main}
    For any employed DA, \ourstrategy which replaces \cref{eq:pospair} by \cref{eq:GPS} in any SSL loss (\cref{eq:SSL}) recovers (i) input space nearest neighbor positive sampling when $g_{\gamma}$ is the identity and $\tau \gg 0$, (ii) standard SSL when $g_{\gamma}$ is a bijection and $\tau \rightarrow 0$, and (iii) NNCLR when $g_{\gamma}=f_{\theta}$ and $\tau \rightarrow 0$.
\end{proposition}

The above result provides a first strong argument demonstrating how \ourstrategy does not reduce the capacity of SSL, in fact, it introduces a novel axis of freedom--namely the design of $(g_{\gamma}, \tau)$--to extend current SSL beyond what is amenable solely by tweaking the original architecture $f_{\theta}$, or the original DA. In particular, the core motivation of the presented method is that this novel ability to design $g_{\gamma}$ also reduces the burden to design DA. In fact, if we consider the case where the original DA is part of the original dataset \begin{align}
    \forall \vx \in \sX, \exists \rho : \text{DA}_{\rho}(\vx) \in \sX,\label{eq:case}
\end{align}
so that for any sample in $\sX$, there exists at least one DA configuration ($\rho$) that produces another sample, \ourstrategy can recover standard SSL albeit without employing any DA at all, as formalized below.

\begin{theorem}
\label{thm:main}
    Performing standard SSL (employing \cref{eq:pospair} into \cref{eq:SSL}) with a given DA and a training set for which \cref{eq:case} holds, is equivalent to performing \ourstrategy (employing \cref{eq:GPS} into \cref{eq:SSL}) without any DA and by setting $g_{\gamma}$ to be invariant only to that DA, i.e. $g_{\gamma}(DA(\vx))=g_{\gamma}(\vx)$.
\end{theorem}

By construction from \cref{eq:case} and assuming that one has the ability to design such an invariant $g_{\gamma}$, it is clear that the nearest neighbour within the training set for any $\vx \in \sX$ will be the corresponding samples $\text{DA}(\vx)$, therefore proving \cref{thm:main}. That result is quite impractical as designing such a mapping $g_{\gamma}$ may prove as arduous as design its underlying DA, but nevertheless provides a great motivation to \ourstrategy. Namely, \ourstrategy does not limit SSL in any way and simply transfer the ability to embed a priori knowledge into SSL from a DA-only position to designing embedding spaces where Euclidean distances become semantically informative. Having the ability to design $g_{\gamma}$ not only has the ability to encompass the burden of design a DA, and both can be used jointly. 

{\bf The design of $g_{\gamma}$.}~
The proposed strategy (\cref{eq:GPS}) is based on finding the nearest neighbors of different candidate inputs in a given embedding space. There are multiple ways for acquiring an informative embedding space, i.e., a prescribed mapping $g_{\gamma}$. Throughout our study, we will focus on the most direct solution of employing a previously pretrained mapping. The pre-training may or may not have occurred on the same dataset being considered for SSL. Naturally, the alignment between both datasets affects the quality and reliability of the embeddings. If one does not have access to such pretrained model, another solution is to first learn an abstracted representation, e.g., an auto-encoder or VAE \citep{kingma2013auto}, and then use the encoder for $g_{\gamma}$. In that setting the motivation lies in the final SSL representation being superior to solve downstream tasks that the encoder ($g_{\gamma}$) alone. We provide some examples of the resulting positive pairs with our strategy in Figure \ref{fig:pknn}. In this figure, we use a pretrained model to calculate the set of \kk nearest neighbors for each image $\vx$ in the target dataset. Then for each image $\vx$, the model randomly chooses the positive image from the nearest neighbors in embedding space (recall \cref{eq:GPS}). Finally, both the original image and the produced positive sample are augmented using the chosen DA and passed as a positive pair of images through the encoders. 
Note that as per \cref{prop:main}, \ourstrategy may choose the image itself as its own positive sample, but the probability of it happening reduces as $\tau$ increases.
As we will demonstrate the later sections, the proposed positive sampling strategy often outperforms the baseline DA based positive pair sampling strategy on multiple datasets.

\begin{figure*}
    \centering
    \includegraphics[width=0.9\textwidth]{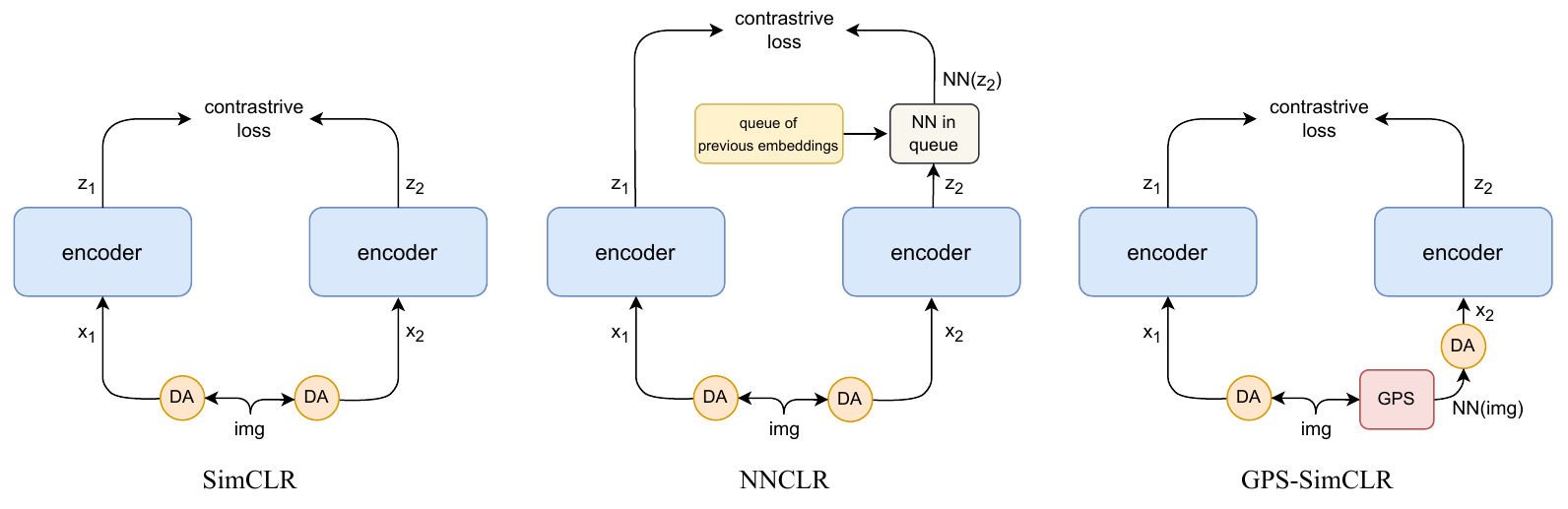}
    \caption{\small Architectures of SimCLR, NNCLR, and \ourmethod. This figure demonstrates where the data augmentaiton (DA) happens in each method and also how the nearest neighbor (NN) search is different between NNCLR and \oursimclr. Note that the `queue' in NNCLR has a limited size, usually set to 65536. This issue could lead to under-represented classes to not be learned efficiently.}
    \label{fig:gnnclr}
\end{figure*}

\paragraph{Relation to NNCLR}
The commonality of NNCLR and \ourstrategy has been brought forward in \cref{prop:main}. In short, they both choose the nearest neighbor of input images as the positive sample. However, the embedding space in which the nearest neighbor is chosen is different; in NNCLR, the model being trained creates the embedding space which is thus updated at every training step, i.e., $g_{\gamma} = f_{\theta}$. However, \ourstrategy generalizes that in the sense that the nearest neighbors can stem from any prescribed mapping, without the constraint that it is trained as part of the SSL training, or even that it takes the form of a DNN. The fact that NNCLR only considers the model being trained to obtain its positive samples also makes it heavily dependent on complex and strong augmentations to produce non degenerate results. On the other hand, our ability to prescribe other mappings for the nearest neighbor search makes \ourstrategy much less tied to the employed DA. We summarize those methods in Figure \ref{fig:gnnclr}.~\looseness-1

\begin{figure*}
    \centering
    \includegraphics[width=1.0\textwidth]{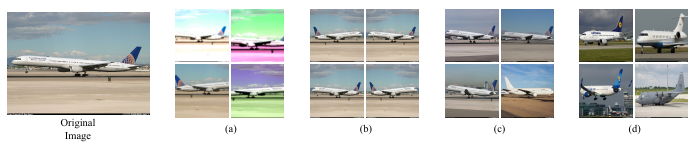}
    \vspace{-0.3cm}
    \caption{\small An example (a) \strongaug and (b) \weakaug applied to an image from the \aircrafts dataset. Furthermore, (c) and (d) depict examples of the 4 nearest neighors calculated by CLIP and VAE embeddings, respectively.~\looseness-1}
    \label{fig:strong-vs-weak}
\end{figure*}

\subsection{Empirical Validation on Benchmarked Datasets} 

In our experiments, we train the baseline SSL methods and the proposed \ourstrategy with two general sets of augmentations, \strongaug, which are augmentations that have been finetuned on the target dataset (for \cifarten \citep{krizhevsky2009learning}) or ImageNet in the case of under-studied datasets (for \aircrafts \citep{maji2013fine}, \pathmnist \citep{medmnistv2}, \tissuemnist  \citep{medmnistv2}, and \hotelsid), and \weakaug, representing the scenario where we do not know the correct augmentations and use minimal ones.
The set of \strongaug consists of \texttt{random-resized-crop}, \texttt{random-horizontal-flip}, \texttt{color-jitter}, \texttt{gray-scale}, \texttt{gaussian-blur} while \texttt{solarization} while \weakaug only uses \texttt{random-horizontal-flip}.

\begin{table}[t!]
\renewcommand{\arraystretch}{0.8}
\centering
\caption{\small Classification accuracy of a ResNet18 in different ablation settings; \textbf{Left:} Comparison \oursimclr when different pretrained networks are used for generating embeddings for nearest-neighbor calculation, i.e., prior knowledge. \textbf{Right:} Best performance in \strongaug setting of SimCLR and \oursimclr given different learning rates (LR). }
\resizebox{0.49\linewidth}{!}{%
    \label{gps-source-compare}    
    \begin{tabular}{l l c l}
    \toprule
     \multirow{2}{*}{\textbf{\edit{\oursimclr}}} & \multicolumn{2}{c}{\aircrafts} \\
     \cmidrule{2-3}
      & \weakaug & \strongaug \\
     \midrule
    \multicolumn{1}{c|}{$\text{ViT-B}_{MAE}$} & 10.53 & 29.55\\
    \multicolumn{1}{c|}{$\text{ViT-L}_{MAE}$} & 14.70 & 35.28\\
    \multicolumn{1}{c|}{$\text{RN50}_{SUP}$} & 18.15 & 41.47\\
    \multicolumn{1}{c|}{$\text{RN50}_{VAE}$} & 11.04 & 32.06\\
    \multicolumn{1}{c|}{$\text{RN50}_{CLIP}$} & \textbf{19.38} & \textbf{50.08}\\
    \bottomrule
    \end{tabular}
}
\resizebox{0.415\linewidth}{!}{%
    \label{lr-comparison}    
    \begin{tabular}{l l c l}
    \toprule
     \multirow{2}{*}{\textbf{LR}} & \multicolumn{2}{c}{\aircrafts} \\
     \cmidrule{2-3}
      & SimCLR & \oursimclr \\
     \midrule
    \multicolumn{1}{c|}{$0.003$} & 21.39	& 35.7\\
    \multicolumn{1}{c|}{$0.01$}  & 30.18	& 43.68	\\
    \multicolumn{1}{c|}{$0.03$}  & 39.27	& 49.57	\\
    \multicolumn{1}{c|}{\edit{$0.1$}}  & \edit{39.81}	& \edit{50.08}	\\
    \multicolumn{1}{c|}{\edit{$0.3$}}  & \edit{39.87}	& \edit{48.10}	\\
    \bottomrule
    \end{tabular}
}
\label{tab:ablation1}
\vspace{-1.5ex}
\end{table}

In order to thoroughly validate \ourstrategy as an all-purpose strategy for SSL, we consider SimCLR, BYOL, NNCLR, and VICReg as baseline SSL models, and for each of them, we will consider the standard SSL positive pair generation (\cref{eq:pospair}) and the proposed one (\cref{eq:GPS}). We opted for a \textit{randomly-initialized} backbone ResNets \citep{he2016deep} as the encoder.
We also bring forward the fact that most SSL methods are generally trained on a large dataset for which strong DAs are known and well-tuned, such as Imagenet, and the learned representation is then transferred to solve tasks on smaller and less known datasets. In many cases, training those SSL models directly on those atypical dataset lead to catastrophic failures, as the optimal DAs have not yet been discovered. Lastly, we will consider five different embeddings for $g_{\gamma}$, one obtained from supervised learning, one from CLIP training \citep{radford2021learning} trained on LAION-400M \cite{schuhmann2021laion}, one for VAE \citep{kingma2013auto}, and two for MAE \citep{he2022masked} all trained on ImageNet and furthermore show our method is more robust to hyper-parameter changes (Table \ref{tab:ablation1}). Before delving in our empirical experiments, we emphasize that the supervised $g_{\gamma}$ is employed  in the \weakaug setting for two reasons. First, it provides what could be thought of as an optimal setting where the class invariants have been learned through the label information. Second, as a mean to demonstrate how one could combine a supervised dataset as a mean to produce prior information into training an SSL model on a different dataset. Since the said models are trained on ImageNet, all the provided results throughout this study remain practical since the labels of the target datasets, on which SSL models are trained and evaluated, are never observed for the training of neither $g_{\gamma}$ nor $f_{\theta}$.

\begin{table*}[t!]
\renewcommand{\arraystretch}{0.8}
\caption{\small Classification accuracy of baseline SSL methods with and without \ourstrategy on four datasets on \edit{\emph{ResNet50}} using pretrained $RN50_{CLIP}$ embeddings for positive sampling. We consider both \strongaug (Strong Augmentation) and \weakaug (Weak Augmentation) settings. The set of DA used for \strongaug are \texttt{random-resized-crop}, \texttt{random-horizontal-flip}, \texttt{color-jitter}, \texttt{gray-scale}, \texttt{gaussian-blur}, and \texttt{solarization}. For the \weakaug setting, the only DA used is \texttt{random horizontal flip}. We mark the \FB{first}, \SB{second}, and \TB{third} best performing models accordingly.}
\begin{center}
\resizebox{\textwidth}{!}{
\begin{tabular}{cccccc}
    \toprule
    \multirow{3}{*}{\textbf{Aug.}} & \multirow{3}{*}{\textbf{Method}} & \multicolumn{4}{c}{\textbf{Datasets}} \\ \cmidrule{3-6}
    & &  \cifarten & \aircrafts & \pathmnist & \tissuemnist \\
    & & \small{(10 classes)} & \small{(100 classes)} & \small{(9 classes)} & \small{(8 classes)} \\
    
    \bottomrule
    \toprule
    \multirow{9}{*}{\rotatebox{90}{\weakaug}} & \multicolumn{1}{c|}{SimCLR} & \edit{47.01} & \edit{5.61} & \edit{63.42} & \edit{50.35} \\
    & \multicolumn{1}{c|}{BYOL} & \edit{41.79} & \edit{6.63} & \edit{67.08} & \edit{48.00} \\
    & \multicolumn{1}{c|}{NNCLR} & \edit{28.46} & \edit{6.33} & \edit{56.70} & \edit{37.98} \\
    & \multicolumn{1}{c|}{Barlow Twins} & \edit{41.73} & \edit{5.34} & \edit{53.27} & \edit{43.57} \\
    & \multicolumn{1}{c|}{VICReg} & \edit{37.51} & \edit{6.18} & \edit{46.46} & \edit{39.79} \\ \cmidrule{2-6}
    & \multicolumn{1}{c|}{\oursimclr (ours)} & \SB{85.08} & \SB{18.18} & \TB{87.79} & \TB{53.14} \\
    & \multicolumn{1}{c|}{\ourbyol (ours)} & \edit{84.07} & \edit{13.50} & \edit{87.67} & \edit{53.05} \\
    & \multicolumn{1}{c|}{\ourbarlow (ours)} & \TB{84.45} & \TB{17.34} & \SB{88.77} & \FB{56.63} \\
    & \multicolumn{1}{c|}{\ourvicreg (ours)} & \FB{85.58} & \FB{18.81} & \FB{88.91} & \SB{56.44} \\ \midrule \midrule
    \multirow{9}{*}{\rotatebox{90}{\strongaug}} & \multicolumn{1}{c|}{SimCLR} & \edit{90.24} & \SB{47.11} & \FB{93.64} & \TB{58.53} \\
    & \multicolumn{1}{c|}{BYOL} & \edit{90.50} & \edit{34.23} & \SB{93.29} & \edit{56.63} \\
    & \multicolumn{1}{c|}{NNCLR} & \edit{90.03} & \edit{34.80} & \edit{92.87} & \edit{52.57} \\
    & \multicolumn{1}{c|}{Barlow Twins} & \edit{88.34} & \edit{18.12} & \edit{92.03} & \FB{61.69} \\
    & \multicolumn{1}{c|}{VICReg} & \FB{91.21} & \edit{38.74} & \TB{93.22} & \SB{60.18} \\ \cmidrule{2-6}
    & \multicolumn{1}{c|}{\oursimclr (ours)} & \SB{91.17} & \FB{55.60} & \edit{92.30} & \edit{55.59} \\
    & \multicolumn{1}{c|}{\ourbyol (ours)} & \TB{91.15} & \edit{44.28} & \edit{92.40} & \edit{55.03} \\
    & \multicolumn{1}{c|}{\ourbarlow (ours)} & \edit{88.52} & \edit{15.47} & \edit{91.98} & \edit{57.04} \\
    & \multicolumn{1}{c|}{\ourvicreg (ours)} & \edit{89.71} & \TB{47.29} & \edit{92.55} & \edit{55.79} \\ \bottomrule
\end{tabular}
}
\end{center}

\label{tab:all-datasets-metrics}
\vspace{-5ex}
\end{table*}

\begin{figure*}
\vspace{-4ex}
    \centering
    \includegraphics[width=0.9\textwidth]{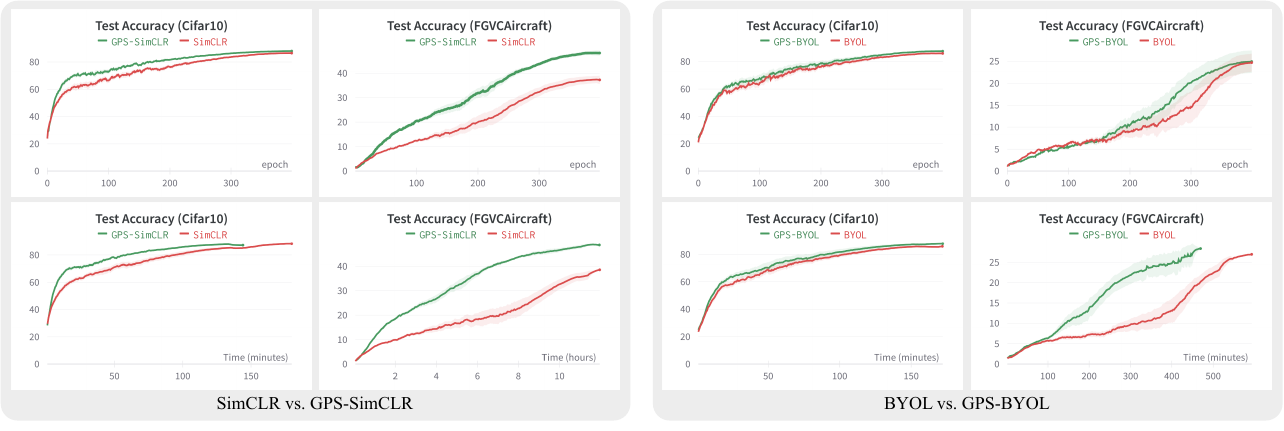}

    \caption{\small Comparing the runtime of BYOL vs. \ourbyol and SimCLR vs. \oursimclr for two datasets, i.e., \aircrafts and \cifarten. In general, we see while the runtime of \ourstrategy remains the same as the original baseline SSL method, it improves the performance.\vspace{-3ex}}
    \label{fig:time-compare}
\end{figure*}

{\bf Strong Augmentation Experiments.}~The DAs in the  \strongaug configuration consist of strong augmentations that usually distort the size, resolution, and color characteristics of the original image. First, we note that in this setting, \ourstrategy generally does not harm the performance of the baseline SSL model on common datasets, i.e. \cifarten (Table \ref{tab:all-datasets-metrics}). In fact, \ourstrategy performs comparable to the best-performing baseline SSL model on \cifarten, i.e., VICReg, showcasing that \ourstrategy does not negatively impact performances even on those datasets. We believe that the main reason lies in the fact that the employed DA has been specifically designed for those datasets (and ImageNet). However, we observe that \ourstrategy outperforms (on \aircrafts and \tissuemnist) or is comparable to (on \pathmnist) the baseline SSL methods for the under-studied and real-word datasets. (Table \ref{tab:all-datasets-metrics}). The reason for this is that, the optimal set and configuration of DA for one dataset is not necessarily the optimal set and configuration for another, and while SSL solely relies on DA for its positive samples, \ourstrategy is able to alleviate that dependency through $g_{\gamma}$ and uses positve samples that can be more useful than default DAs, as seen in Figure \ref{fig:strong-vs-weak}. The results for these experiments can be seen in Table \ref{tab:all-datasets-metrics}. Note that our method's runtime is similar to the baseline SSL method on the dataset it is learning and does not hinder the training process (Figure \ref{fig:time-compare}).


{\bf Weak-Augmentation Experiments.}~We perform all experiments under the \weakaug setting as well, showing \ourstrategy produces high quality representations even in that setting, validating \cref{thm:main}. As seen in Table \ref{tab:all-datasets-metrics}, \ourstrategy significantly outperforms all baseline SSL methods across both well-studied and under-studied datasets. 
These results show that our \ourstrategy strategy, though conceptually simple, coupled with the \weakaug setting, approximates strong augmentations used in the \strongaug configuration. This creates a significant advantage for \ourstrategy to be applied to real-world datasets where strong augmentations have not been found, but where the invariances learned by $g_{\gamma}$ to generalize to them.

{\bf Ablation Study} In this section we explore multiple ablation experiments in order to show \ourstrategy improves SSL and is indeed a future direction for improving SSL methods. \edit{First, we compare SSL and GPS training on \cifarten and \aircrafts starting from a backbone initialized with random (realistic setting), supervised ImageNet pretrained, or CLIP pretrained weights to explore whether the improvement of \ourstrategy is due to better positive sampling or simply because of using a strong prior knowledge. We show in Table \ref{tab:scratch-vs-pretrained} that \ourstrategy performs better than the baseline SSL methods, even when they both have access to the pretrained network weights. This proves that the improvement in performance of \ourstrategy compared to baseline SSL methods is indeed due to better positive sampling.}~\looseness-1

Next, we compare \oursimclr with three different embeddings for $g_{\gamma}$; supervised, VAE, and CLIP embeddings. We observe that as the embeddings get higher in quality based on the pre-trained network, as the performance increases in both the \weakaug and \strongaug setting. However, note that even given the worst embeddings, i.e., the VAE embeddings, \oursimclr still outperforms the original SimCLR in the \weakaug setting, showcasing that the nearest neighbors add value to the learning process when the augmentations are unknown.~\looseness-1

\edit{We further explore if the improvement of \ourstrategy holds when methods are trained longer. To that end, we train a ResNet18 for 1000 epochs with SimCLR and VICReg with \strongaug, along with their GPS versions, on \cifarten and \aircrafts and compare the results with the performance from 400 epochs. As seen in Table \ref{tab:longer-epochs}, the improvement of \ourstrategy compared to the baseline SSL method holds on \aircrafts dataset and remains comparable on \cifarten, showcasing the robustness of \ourstrategy.}~\looseness-1

\begin{table}
\renewcommand{\arraystretch}{0.8}
\caption{\small \edit{Test accuracy comparison of \ourstrategy after 1K epochs versus 400 training epochs. We show the improvements of \ourmethod are still significant on \aircrafts and comparable on \cifarten.~\looseness-1}}

\begin{center}
\resizebox{0.65\textwidth}{!}{
\begin{tabular}{ccccc}
\toprule
 \multirow{2}{*}{\textbf{Method}}& \multicolumn{2}{c}{\textbf{\cifarten}} & \multicolumn{2}{c}{\textbf{\aircrafts}} \\
\cmidrule{2-5}
 & \edit{400 eps} & \edit{1000 eps} & \edit{400 eps} & \edit{1000 eps} \\
\bottomrule
\toprule
\edit{SimCLR} & \edit{88.26} & \textbf{\edit{91.25}} & \edit{39.87} & \edit{45.55} \\
\edit{\oursimclr} & \textbf{\edit{89.57}} & \edit{91.10} & \textbf{\edit{50.08}} & \textbf{\edit{51.64}} \\
\midrule
\edit{VICReg} & \edit{89.34} & \textbf{\edit{90.61}} & \edit{33.21} & \edit{41.19} \\
\edit{\ourvicreg} & \textbf{\edit{89.68}} & \edit{89.84} & \textbf{\edit{45.48}} & \textbf{\edit{49.29}} \\
\bottomrule
\end{tabular}

}
\end{center}
\label{tab:longer-epochs}
\vspace{-4.5ex}
\end{table}

Finally, we aim to measure the sensitivity of the performance of a baseline SSL method to a hyper-parameter, i.e., learning rate, with and without \ourstrategy. In this ablation experiment, we report the best performance of SimCLR and \oursimclr given different learning rates in the \strongaug setting. We observe that \ourstrategy when applied to a baseline SSL method is as much, if not more, robust to hyper-parameter change. The results of both ablations are reported in Table \ref{tab:ablation1}. We further compare \ourstrategy with linear probing's performance and other ablations in Appendix \ref{sec:ablation}.~\looseness-1

\begin{table*}[ht!]
\renewcommand{\arraystretch}{0.8}
\caption{\small \edit{Comparing SimCLR with and without \ourmethod with different initializations with a ResNet50. RAND, $\text{PT}_{SUP}$, and $\text{PT}_{CLIP}$ represent random weights, ImageNet supervised weights, and CLIP pretrained weights.}}
\begin{center}
\resizebox{0.9\textwidth}{!}{
\begin{tabular}{cccccc}
\toprule
\multirow{2}{*}{\textbf{Method}} & \multirow{2}{*}{\textbf{Weight Init.}} & \multicolumn{2}{c}{\textbf{\cifarten}} & \multicolumn{2}{c}{\textbf{\aircrafts}} \\  \cmidrule{3-6}
& & \textbf{Weak Aug} & \textbf{Strong Aug} & \textbf{Weak Aug} & \textbf{Strong Aug} \\
\bottomrule
\toprule
\edit{SimCLR} & \multirow{2}{*}{\edit{RAND}} & \edit{46.69} & \edit{87.39} & \edit{5.67} & \edit{27.36} \\
\edit{\oursimclr} & & \textbf{\edit{85.2}} & \textbf{\edit{90.48}} & \textbf{\edit{17.91}} & \textbf{\edit{43.56}} \\ \midrule
\edit{SimCLR} & \multirow{2}{*}{\edit{$PT_{SUP}$}} & \edit{43.99} & \edit{94.02} & \edit{17.91} & \edit{59.92} \\
\edit{\oursimclr} & & \textbf{\edit{91.3}} & \textbf{\edit{95.53}} & \textbf{\edit{39.45}} & \textbf{\edit{66.88}} \\ \midrule
\edit{SimCLR} & \multirow{2}{*}{\edit{$PT_{CLIP}$}}  & \edit{45.57} & \edit{90.26} & \edit{6.21} & \edit{41.04} \\
\edit{\oursimclr} & & \textbf{\edit{89.44}} & \textbf{\edit{91.23}} & \textbf{\edit{24.15}} & \textbf{\edit{49.63}} \\
\bottomrule
\end{tabular}
}
\end{center}
\label{tab:scratch-vs-pretrained}
\vspace{-2.5ex}
\end{table*}



\vspace{-2ex}
\section{Case Study on the Hotels Image Dataset}
\vspace{-2ex}
\label{sec:dataset}

In this section, we study how \ourstrategy compares to baseline SSL methods on an under-studied real-world dataset. We opt the \hotelsid \cite{} dataset for our evaluation which gathers hotel images for the purpose of countering human-trafficking. \hotelsid provides a single train set alongside 4 evaluation sets, each with a different level of difficulty.~\looseness-1

\begin{table}[ht!]
\renewcommand{\arraystretch}{0.8}
\caption{\small \recallatone on different spltis on \hotelsid Dataset for SSL methods. The splits are namely, \firsttest: \{branch: seen, chain: seen\}, \secondtest: \{branch: unseen, chain: seen\}, \thirdtest: \{branch: unseen, chain: unseen\} and  \fourthtest: \{branch: unknown, chain: unknown\}.  We mark the best performing score in \textbf{bold}.}
\begin{center}
    {%
        \begin{tabular}{clllll}
            \toprule
            \multicolumn{1}{c}{\textbf{Method}}   & \firsttest            &      \secondtest           &   {\thirdtest}             &       \multicolumn{1}{c}{\fourthtest}      \\
            \bottomrule
            \toprule
            \multicolumn{1}{c}{{SimCLR}}    &   \multicolumn{1}{|c}{3.28} &        16.76         &       20.30         &       \multicolumn{1}{c}{16.00}    \\

            \multicolumn{1}{c}{{BYOL}}           &   \multicolumn{1}{|c}{3.69}                       &       19.27                           &       23.02      &       \multicolumn{1}{c}{18.47}   \\           

            \multicolumn{1}{c}{{Barlow Twins}}         &   \multicolumn{1}{|c}{3.04}    &       15.54               &       18.96            &       \multicolumn{1}{c}{15.06}   \\

            \multicolumn{1}{c}{{VICReg}}   &   \multicolumn{1}{|c}{3.41}        &       17.52                  &        20.45            &       \multicolumn{1}{c}{16.53}  \\        

            \midrule
            
            \multicolumn{1}{c}{$\oursimclr$ (ours)}    &   \multicolumn{1}{|c}{\SB{4.84}}                 &        \SB{23.67}                 &       \SB{26.30}                &       \multicolumn{1}{c}{\SB{22.28}}  \\
            \multicolumn{1}{c}{$\ourbyol$ (ours)}    &   \multicolumn{1}{|c}{3.89}                 &        19.64                 &       23.18                &       \multicolumn{1}{c}{19.38}  \\
            \multicolumn{1}{c}{$\ourbarlow$ (ours)}    &   \multicolumn{1}{|c}{\TB{4.49}}                 &        \TB{21.98}                 &       \TB{25.23}                &       \multicolumn{1}{c}{\TB{20.82}}  \\
            \multicolumn{1}{c}{$\ourvicreg$ (ours)}    &   \multicolumn{1}{|c}{\FB{5.33}}                 &        \FB{25.71}                 &       \FB{28.29}                &       \multicolumn{1}{c}{\FB{23.78}}  \\            
            \bottomrule
        \end{tabular}
\vspace{-2ex}
    }
\end{center}
\label{tab:hotelid-ssl-metrics}
\end{table}

We evaluate the baseline SSL models with and without \ourstrategy to the \hotelsid dataset and report the Recall@1 (\recallatone) for the different splits introduced. Based on the findings from \ref{tab:all-datasets-metrics}, we adapt the \strongaug setting along with the prior knowledge generated by a CLIP-pretrained ResNet50.~\looseness-1

As seen in Table \ref{tab:hotelid-ssl-metrics}, SSL baselines always get an improvement when used with \ourstrategy. The reason the baseline SSL methods underperform compared to their \ourstrategy version is that the positive samples generated only using DA lack enough diversity since the images from \hotelsid dataset have various features and merely DAs limits the information the network learns; however, paired with \ourstrategy, we see a clear boost in performance across all different splits due to the additional information added by the neareset neighbors.~\looseness-1

\section{Conclusions}
\label{sec:concl}
In this paper we proposed \ourstrategy which presents a novel strategy to obtain positive samples for Self-Supervised Learning. In particular, \ourstrategy moves away from the usual DA-based positive sampling by instead producing positive samples from the nearest neighbors of the data as measure in some prescribed embedding space. That is, \ourstrategy introduces an entirely novel axis to research and improve SSL that is complementary to the design of DA and losses. Through that strategy, we were for example able to train SSL on atypical datasets such as medical images --without having to search and tune for the right DA. Those results open new avenues to the existing strategy of SSL pretraining on large dataset, and then transferring the model to those other datasets for which DAs as not available. In fact, we observe that while \ourstrategy meets or surpass SSL performances across our experiments, the performance gap is more significant when the optimal DAs are not known, e.g., in \pathmnist and \tissuemnist we observe the performance of \ourstrategy with weak augmentations is slightly less than with strong augmentations. Besides practical applications, \ourstrategy finally provides a novel strategy to embed prior knowledge into SSL.~\looseness-1

{\bf Limitations.}~The main limitation of our method is akin to the one of SSL, it requires the knowledge of the embedding in which the nearest neighbors are obtain to produce the positive samples. This limitation is on par with standard SSL's reliance on DA, but its formulation is somewhat dual (recall \cref{thm:main}) in that one may know how to design such an embedding without knowing the appropriate DA for the dataset, and vice-versa. Alternative techniques like training separate and simple DNs to provide such embeddings prior to the SSL learning could be considered for future research. ~\looseness-1

\bibliographystyle{iclr2024_conference}
\bibliography{iclr2024_conference}


\appendix
\section{Appendix}
\subsection{\hotelsid Splitting method}
\label{app:splitting-method}
\edit{\hotelsid \citep{feizi2022revisiting} is created carefully to make sure no data leakage occurs. They mention how the total data is divided into the train and the multiple test splits. More specifically, first a set of chains (along with \textit{all} their branches) are reserved for the \thirdtest to make sure the chains (super-classes) and branches (classes) are not seen during training. Next, out of the remaining chains, a set of the branches are chosen to add \textit{all} of their images to the \secondtest test split (since the training set will have other images from other branches from the same chain, but not the same branch images). Finally, out of the remaining branches, the images in each are split between \firsttest and train, creating the final test split that has a subset of the branches seen during training. With this procedure, they make sure of the table of overlapping below. More details regarding the splits is provided in the original paper.}
\subsection{Hyper-Parameter Search}

In all experiments, we train for \edit{400 epochs} with a batch size of 256 using one RTX 8000 GPU for all methods. To ensure we are choosing the correct hyper-parameters for a fair comparison, we search over a vast range of hyper-parameter combinations ($lr \in \{1e^{-3}, 3e^{-3}, 3e^{-2}, 1e^{-2}, 3e^{-1}, 1e^{-1}, 1\}$, $classifier\_lr \in \{3e^{-2}, 1e^{-2}, 3e^{-1}, 1e^{-1}, 1, 3\}$,  $weight\_decay \in \{1e^{-4}, 1e^{-3}\}$) and for \ourstrategy with all SSL baselines we also search over $k \in \{1, 4, 9, 49\}$). \edit{For experiments using \weakaug and \strongaug, we use nearest neighbors calculated based on embeddings created from a ResNet50 that have been CLIP pre-trained as the prior knowledge.} Finally, for each method, we report the best classification accuracy for \cifarten, \aircrafts, \pathmnist, and \tissuemnist, and Recall@1 (\recallatone) for \hotelsid in Tables \ref{tab:all-datasets-metrics}, \ref{tab:hotelid-ssl-metrics}, and \ref{tab:hotelid-ssl-pretrained}. To calculate both metrics, we first train the encoder on the target dataset using the SSL method, with or without \ourstrategy. Then, for classification accuracy, we train a linear classifier on top of it, and for \recallatone, we encode all the images from the test set and calculate the percentage of images which their first nearest neighbor is from the same class.

\subsection{Ablation Study}
\label{sec:ablation}

\subsubsection{Different Backbone}
\label{sec:diff-backbone-ablation}
\edit{First, we provide the same experiments as in Table \ref{tab:all-datasets-metrics}, but trained with a ResNet18 instead of a ResNet50 and provide the results in Table \ref{tab:all-datasets-metrics-rn18}. We see the same results for ResNet50 (discussed for Table \ref{tab:all-datasets-metrics}) also hold when ran on a smaller architecture, i.e., ResNet18. This shows the improvements of \ourstrategy over baseline SSL methods is more reliable and robust.}

\begin{table*}[t]
\caption{Classification accuracy of baseline SSL methods with and without \ourstrategy on four datasets on \emph{ResNet18} using pretrained $RN50_{CLIP}$ embeddings for positive sampling.. We consider both \strongaug (Strong Augmentation) and \weakaug (Weak Augmentation) settings. The set of DA used for \strongaug are \texttt{random-resized-crop}, \texttt{random-horizontal-flip}, \texttt{color-jitter}, \texttt{gray-scale}, \texttt{gaussian-blur}, and \texttt{solarization}. For the \weakaug setting, the only DA used is \texttt{random horizontal flip}. We mark the \FB{first}, \SB{second}, and \TB{third} best performing models accordingly.}
\begin{center}
\resizebox{\textwidth}{!}{
\begin{tabular}{ccccccc}
    \toprule
     \multirow{3}{*}{\textbf{Aug.}} & \multirow{3}{*}{\textbf{Method}} & \multicolumn{4}{c}{\textbf{Datasets}} \\ \cmidrule{3-6}
    &  & \cifarten & \aircrafts & \pathmnist & \multicolumn{1}{c}{\tissuemnist} \\
    &  & \small{(10 classes)} & \small{(100 classes)} & \small{(9 classes)} & \small{(8 classes)} \\
    
    \bottomrule
    \toprule
    \multirow{9}{*}{\rotatebox{90}{\weakaug}} &  \multicolumn{1}{c|}{SimCLR}  &   \edit{47.62} & \edit{7.70} & \edit{62.99} & \multicolumn{1}{c}{\edit{52.30}} \\
    & \multicolumn{1}{c|}{BYOL} & \edit{49.72} & \edit{8.99} & \edit{77.77} & \multicolumn{1}{c}{\edit{51.00}} \\
    & \multicolumn{1}{c|}{NNCLR} & \edit{71.74} & \edit{8.10} & \edit{56.92} & \multicolumn{1}{c}{\edit{42.59}} \\
    & \multicolumn{1}{c|}{Barlow Twins} & \edit{42.00} & \edit{7.53} & \edit{64.82} & \multicolumn{1}{c}{\edit{49.43}} \\
    & \multicolumn{1}{c|}{VICReg} & \edit{36.04} & \edit{4.95} & \edit{56.92} & \multicolumn{1}{c}{\edit{50.26}} \\ \cmidrule{2-6} 
    &  \multicolumn{1}{c|}{\oursimclr (ours)} & \FB{85.83} & \SB{18.48} & \FB{88.62} & \multicolumn{1}{c}{\SB{55.98}} \\
    & \multicolumn{1}{c|}{\ourbyol (ours)} & \edit{84.56} & \edit{14.79} & \edit{81.66} & \multicolumn{1}{c}{\FB{56.21}} \\
    & \multicolumn{1}{c|}{\ourbarlow (ours)} & \TB{84.83} & \TB{18.12} & \TB{87.79} & \multicolumn{1}{c}{\TB{55.86}} \\
    & \multicolumn{1}{c|}{\ourvicreg (ours)} & \SB{85.38} & \FB{20.16} & \SB{87.83} & \multicolumn{1}{c}{\edit{55.26}} \\ \midrule \midrule
    \multirow{9}{*}{\rotatebox{90}{\strongaug}} & \multicolumn{1}{c|}{SimCLR} & 88.26 & \TB{39.87} & 91.56 & \multicolumn{1}{c}{61.51} \\
    & \multicolumn{1}{c|}{BYOL} & 86.90 & 27.33 & 91.24 & \multicolumn{1}{c}{60.73} \\
    & \multicolumn{1}{c|}{NNCLR} & 87.95 & 39.12 & 91.14 & \multicolumn{1}{c}{52.42} \\
    & \multicolumn{1}{c|}{Barlow Twins} & 88.89 & 25.71 & \SB{92.23} & \multicolumn{1}{c}{60.06} \\
    & \multicolumn{1}{c|}{VICReg} & \TB{89.34} & 33.21 & \FB{92.27} & \multicolumn{1}{c}{59.41} \\ \cmidrule{2-6}
    & \multicolumn{1}{c|}{\oursimclr (ours)} & \SB{89.57} & \FB{50.08} & \TB{92.19} & \multicolumn{1}{c}{\SB{62.76}} \\
    & \multicolumn{1}{c|}{\ourbyol (ours)} & 88.46 & 32.07 & 91.05 & \multicolumn{1}{c}{54.05} \\
    & \multicolumn{1}{c|}{\ourbarlow (ours)} & 88.39 & 25.35 & 91.55 & \multicolumn{1}{c}{\FB{62.93}} \\
    & \multicolumn{1}{c|}{\ourvicreg (ours)} & \FB{89.68} & \SB{45.48} & 91.88 & \multicolumn{1}{c}{\TB{62.46}} \\ \bottomrule
    
\end{tabular}
}
\end{center}

\label{tab:all-datasets-metrics-rn18}
\end{table*}

\subsubsection{Finetuning for \hotelsid}
\label{sec:finetuning-ablation}
We further try a trivial way of transferring knowledge from a pretrained network to other SSL baseline models and compare it to \ourmethod; we initialize the base encoder in any SSL method, i.e., the ResNet18, to the pretrained network's weights, as opposed to random initialization, and train it i.e., finetuning. Ultimately, we compare the results on \hotelsid in Table \ref{tab:hotelid-ssl-pretrained}.

Although this might perform better if the pretrained network was trained on a visually similar dataset to the target dataset, Table \ref{tab:hotelid-ssl-pretrained} shows that it may harm the generalization on datasets that are different, e.g., ImageNet and \hotelsid, compared to being trained from scratch. However, \ourstrategy proves to be a stable method for transferring knowledge even if the pretrained and target dataset are visually different (Table \ref{tab:hotelid-ssl-metrics}).

\begin{table}[ht!]
\caption{Comparing the \recallatone performance of SSL methods on \hotelsid when trained from scratch against being initialzied to a ImageNet pretrained network. The models with pretrained-initialized encoders (finetuned) are marked with `FT-'. We highlight the difference in \recallatone of the pretrained against the scratch version with {\color{mygreen}{green}} when it improves and {\color{myred}{red}} when it worsens.}
\begin{center}
    \resizebox{0.6\linewidth}{!}{%
        \begin{tabular}{clllll}
            \toprule
            \multicolumn{1}{c}{Method}    & \firsttest            &      \secondtest           &   {\thirdtest}             &       \multicolumn{1}{c}{\fourthtest}      \\
            \bottomrule
            \toprule
            \multicolumn{1}{c|}{{SimCLR}}  &   3.23  &        16.10         &       19.62         &       \multicolumn{1}{c}{15.12}    \\

            \multicolumn{1}{c|}{{FT-SimCLR}}  &   \color{myred}{-0.10}  &      \color{myred}{-0.21}          &       \color{myred}{-0.40}       &       \multicolumn{1}{c}{\color{mygreen}{+0.27}}  \\    
        
            \midrule
            
            \multicolumn{1}{c|}{{BYOL}}           &   3.27                       &       16.25                           &       20.20      &       \multicolumn{1}{c}{15.91}   \\           
            \multicolumn{1}{c|}{{FT-BYOL}}           &   \color{myred}{-0.57}                       &       \color{myred}{-1.75}                           &       \color{myred}{-2.23}   &       \multicolumn{1}{c}{\color{myred}{-1.50}}   \\         
            \midrule
            
            \multicolumn{1}{c|}{{NNCLR}}         &   2.84    &       13.91               &       17.15            &       \multicolumn{1}{c}{13.96}   \\    
            \multicolumn{1}{c|}{{FT-NNCLR}}           &   \color{myred}{-0.54}    &       \color{myred}{-2.44}               &       \color{myred}{-3.18}            &       \multicolumn{1}{c}{\color{myred}{-2.67}}   \\     
            
            \midrule
            
            \multicolumn{1}{c|}{{VICReg}}   &   3.24         &       16.67                  &        19.97            &       \multicolumn{1}{c}{15.86}  \\        
            \multicolumn{1}{c|}{{FT-VICReg}}   &   \color{myred}{-0.43}    &       \color{myred}{-1.54}                 &        \color{myred}{-2.45}                &       \multicolumn{1}{c}{\color{myred}{-1.92}}  \\      
            \bottomrule
            
        \end{tabular}
    }
\end{center}
\vspace{-1ex}
\label{tab:hotelid-ssl-pretrained}

\end{table}

\subsubsection{Comparing to Linear Probing}
\label{sec:lp-ablation}
\edit{Finally, we compare the linear probing performance of the embeddings generated from different architectures, i.e. GPS backbones (GPS-BB), pretrained on different datasets, i.e., GPS Datasets (GPS-DS), with the performance of \ourstrategy using them. More specifically, in Tables \ref{tab:cifar10-lp} and \ref{tab:aircrafts-lp}, we compare the linear probe performance of the CLIP pretrained ResNet50 on LAION-400M \citep{schuhmann2021laion} along with vision transformers (ViTs) pretrained on ImageNet using Masked Auto Encoders (MAE) \citep{he2022masked}, a popular self-supervised method that also does not rely on strong augmentations. We see our method outperforms the linear probe accuracy of CLIP embeddings for both \cifarten and \aircrafts and matches that of ViT-Base and ViT-Large for \cifarten and ViT-Large for \aircrafts.}

\edit{However, we further see that if we train the ViT-Large on the \aircrafts, using MAE with minimal augmentations, we can use that as the positive sampler for \ourstrategy and beat the baseline SSL method on \aircrafts. This shows that \ourstrategy does not entirely rely on huge pretrained models and that there is potential possibilities for training a positive sampler prior to applying \ourstrategy to further boost the performance of baseline SSL methods. 
}

\begin{table}[t!]
\renewcommand{\arraystretch}{0.8}
\caption{\small \edit{Comparison of Linear probing (LP) and \ourvicreg's  (with ResNet50) classification accuracy on \aircrafts with different GPS backbones (GPS BB) pretrained with CLIP and masked auto encoders (MAE) on different datasets without supervision (GPS DS).
The performance of the vanilla VICReg is also depicted for comparison. RN50 and ViT-L refer to ResNet50 and ViT-Large, respectively.}}
\begin{center}
\resizebox{0.8\textwidth}{!}{
\begin{tabular}{cccc|c}
\toprule
 \textbf{GPS-BB} & \textbf{GPS-DS} & \textbf{LP} & \textbf{\ourvicreg} & \textbf{VICReg} \\
\bottomrule
\toprule

$\text{RN50}_{CLIP}$ & \edit{LAION-400M} & \edit{44.55} & \edit{46.44} & \multirow{3}{*}{\edit{39.99}} \\
$\text{ViT-L}_{MAE}$ & \edit{ImageNet} &   \edit{37.32} & \edit{38.44} &  \\
$\text{ViT-L}_{MAE}$ & \edit{\aircrafts}  & \edit{17.01} & \edit{42.87} &  \\
\bottomrule

\end{tabular}
}
\end{center}
\label{tab:aircrafts-lp}
\end{table}

\begin{table}[t!]
\renewcommand{\arraystretch}{0.9}
\caption{\small \edit{Classification accuracy comparison of linear probing (LP) using embeddings with different GPS backbones (GPS-BB) pretrained with CLIP and masked auto encoders (MAE) on different upstream datasets, i.e. GPS-DS, and a trained ResNet50 with \oursimclr on \aircrafts and \cifarten using the same GPS backbones and datasets.
RN50, ViT-L, and Vit-B refer to ResNet50, ViT-Large, and ViT-Base respectively.}}
\begin{center}
\resizebox{0.8\textwidth}{!}{

\begin{tabular}{c c c c c c}
\toprule
 \multirow{2}{*}{\textbf{GPS-BB}} & \multirow{2}{*}{\textbf{GPS-DS}} & \multicolumn{2}{c}{\textbf{\cifarten}} & \multicolumn{2}{c}{\textbf{\aircrafts}} \\
 \cmidrule{3-6}
 & & LP & \edit{\oursimclr} & LP & \edit{\oursimclr} \\
 \bottomrule
 \toprule
$\text{RN50}_{CLIP}$ & \edit{LAION-400M} & \edit{87.85} & {\edit{91.17}} & \edit{44.55} & {\edit{53.81}} \\
$\text{ViT-B}_{MAE}$ & \edit{ImageNet} & \edit{85.78} & {\edit{87.35}} & \edit{27.96} & {\edit{29.55}} \\
$\text{ViT-L}_{MAE}$ & \edit{ImageNet} & {\edit{91.45}} & \edit{90.11} & {\edit{37.29}} & \edit{35.28} \\
$\text{ViT-L}_{MAE}$ & \edit{\aircrafts} & {\edit{-----}} & \edit{-----} & {\edit{17.01}} & \edit{46.93} \\
\bottomrule
\end{tabular}

}
\end{center}
\label{tab:cifar10-lp}
\end{table}

\end{document}